%% file: paper.tex
\documentclass[runningheads]{llncs}
\usepackage[T1]{fontenc}
\usepackage{graphicx}

\usepackage{booktabs}
\usepackage[misc]{ifsym}
\newcommand{\corr}{(\Letter)}

\usepackage{hyperref}
\usepackage{color}

\usepackage{subfigure}
\usepackage{xcolor}
\usepackage{colortbl}
\usepackage{array}
\usepackage{comment}

\definecolor{blackcell}{HTML}{000004}
\definecolor{orangecell}{HTML}{f98e09}
\definecolor{redcell}{HTML}{bc3754}
\definecolor{greencell}{HTML}{249d53}
\definecolor{yellowcell}{HTML}{fcffa4}
\definecolor{lighttext}{HTML}{f1f1f1}
\definecolor{lightorangecell}{HTML}{fbb61a}
\definecolor{darkorangecell}{HTML}{ed6925}

\begin{document}


\title{Evaluating Theory of Mind in Reasoning Models: Robustness over Reasoning}



\author{Ian B. de Haan\inst{1}\orcidID{0009-0009-8074-8265} \and Peter van der Putten\inst{1}\orcidID{0000-0002-6507-6896} \corr \and Max van Duijn\inst{1}\orcidID{0000-0003-0798-9598}}


\authorrunning{I.B. de Haan et al.}


\institute{LIACS, Leiden University, The Netherlands \email{i.b.de.haan.2@umail.leidenuniv.nl},\email{p.w.h.van.der.putten@liacs.leidenuniv.nl, m.j.van.duijn@liacs.leidenuniv.nl}}


\maketitle              

\begin{abstract}
Large language models (LLMs) have recently shown strong performance on Theory of Mind (ToM) tests, prompting debate about the nature and validity of the underlying capabilities. At the same time, reasoning-oriented LLMs trained via reinforcement learning with verifiable rewards have demonstrated notable improvements across a range of benchmarks. In this work, we examine the behavior of such reasoning models in ToM tasks using novel adaptations of machine psychological experiments together with results from established benchmarks. We observe that reasoning models consistently exhibit increased robustness to prompt variations and task perturbations. Our analysis suggests these gains come at least partly from models being more robust at reaching the correct answer under prompt and task variation. We read this as evidence for a robustness-based account rather than for a new ToM-specific ability.
\end{abstract}

\section{Introduction}
In recent years, large language models (LLMs) have become sufficiently capable to be part of our daily lives, and are increasingly used in social and agentic settings \cite{plaat_agentic_2025}. Users tend to attribute human-like intentionality and reasoning to those models, even though the extent to which the models are capable of complex reasoning remains, to a large extent, an open research problem \cite{zhaoSurveyLargeLanguage2025,rahwanMachineBehaviour2019,van-dijk-etal-2023-large,plaat_reasoning_2025}.

Due to the complexity of these systems, even if all information about their architecture is known, it is still difficult to determine and predict their actual behavior. Considering this, researchers in a multitude of fields started investigating those systems and their intelligence by interacting with them and understanding their behavior, rather than by analyzing their architecture.  \cite{rahwanMachineBehaviour2019} 

In this context, previous work has debated the extent to which LLMs demonstrate Theory of Mind (ToM) behavior, that is, the ability to reason about mental states, beliefs, intentions, and desires. Some early claims were optimistic of such a skill emerging to some extent \cite{kosinski_evaluating_2024}. However, follow-up criticism \cite{ullmanLargeLanguageModels2023} led to the creation of benchmarks testing these abilities more comprehensively \cite{vandermeulen2025properly,duijnTheoryMindLarge2023,kimFANToMBenchmarkStresstesting2023,chen2024tombench,hu2025re-evaluatingToM}.

Since then, so-called reasoning model families, like OpenAI's GPT-5, Anthropic's Claude, and the DeepSeek R series, have been released. The innovation behind these models is that they are trained, via reinforcement learning, to ``think before answering'', that is, they are trained to produce a complex chain of thought before answering the query, a form of inference-time scaling \cite{ReasoningModelsOpenAI,paliotta_thinking_2025,Guo2025DeepSeekR1-nature,plaat_reasoning_2025}. This new type of model might not only perform significantly better in ToM tasks, but the chain-of-thought could, in theory, yield important insights about what those models are actually doing to answer the queries.

However, it remains unclear whether these improvements reflect fundamentally stronger ToM reasoning capabilities, or instead improved robustness in finding correct solutions under prompt and task variations. 
Understanding this distinction is important for interpreting apparent progress in social-cognitive abilities of LLMs and for designing reliable evaluation protocols.

This paper studies the role of reasoning in the performance of large language models on ToM tasks, and can be seen as a contribution of humanities and social sciences to AI, and vice versa, though we are only exploring LLM behavior and explicitly not making any strong AI claims about whether LLMs truly possess ToM. First, we evaluate recent reasoning-oriented models using established psychological ToM tests from the battery of \cite{duijnTheoryMindLarge2023}. Second, we introduce prompt variants inspired by \cite{ullmanLargeLanguageModels2023} designed to probe robustness under task-preserving prompt perturbations. Third, we compare reasoning-enabled and non-reasoning model behavior across these tests to examine how reasoning affects performance stability. Finally, we complement our experiments with benchmark results reported in \cite{kimHypothesisDrivenTheoryofMindReasoning2025}. All prompts, data, and code are publicly available at \href{https://github.com/ianbdehaan/Evaluating_ToM_in_Reasoning_Models}{this GitHub repository}, including a supplementary materials document. 

The remainder of this paper is structured as follows: Section \ref{Background} provides an overview of related work on the subject. Section~\ref{methods} introduces the materials and methods employed in the experiments, and Section~\ref{Results} presents the experimental results; Section~\ref{Discussion} interprets the obtained results, points out some limitations of the current work, and suggests new directions in which the research can be expanded; Section~\ref{conclusions} sums up the findings.

\section{Background}\label{Background}

In this section, a bibliographical review in the context of the research done is presented. The contents covered are Section~\ref{CoT and Reas}, a review of what chains of thought are and the recent reinforcement learning with verifiable rewards (RLVR) trained reasoning models, and Section~\ref{ToM and LLMs}, an introduction on the topic of ToM, how it would be a useful skill for LLMs to have, and in Section~\ref{benchResults} third-party benchmarks investigating to which extent they do.

\subsection{Chain of Thought and Reasoning Models} \label{CoT and Reas}

RLVR proved itself to be effective at inducing complex reasoning in LLMs and, using this technique, a new wave of models, refined to produce chains of thoughts together with the answer, started emerging and were named \emph{reasoning models} \cite{xuLargeReasoningModels2025,plaat_reasoning_2025,plaat_agentic_2025}. The implementation details of those models are an important topic, but outside of the scope of this paper.

However, although promising results in several fields, particularly in mathematics and programming, ignited hope that RLVR-trained LLMs could achieve a more general form of intelligence, \cite{yueDoesReinforcementLearning2025} demonstrated that, in reality, the gained performance of those models comes from a more efficient sampling of answers on problems that are already solvable by the base model, not introducing fundamentally new reasoning capabilities. Therefore, the reasoning capacity of a reasoning model remains bounded by its base model.
Reasoning was also initially viewed as promising in the way towards tracking reward hacking in LLMs, but it has been shown that, at least in scenarios where the CoT is not necessary for such behavior, the verbalization of it is very low \cite{chenReasoningModelsDont2025}. This finding also makes one conclude, more generally, that reasoning models don't always include their true reasoning in their think tokens, although the same article also showed that, at least part of the time, they do.

\subsection{Theory of Mind and LLMs} \label{ToM and LLMs}

ToM is generally defined as the ability of an individual to attribute mental states to others (including oneself) and keep track of them, and also tightly relate it to the notion of self and others \cite{baron-cohen_theory_2001,duijnTheoryMindLarge2023}, additional definitions in supplementary materials in the repository.
According to \cite{baron-cohen_does_1985}, “The ability to make inferences about what other people believe to be the case in a given situation allows one to predict what they will do”, this notion not only shows the importance that ToM has in human social interactions, but also allows one to design experiments to test whether an agent demonstrates ToM behavior or not. Quesque and Rossetti \cite{quesque_what_2020} suggest that, for a particular task to act as a valid ToM assessment, it should follow two criteria: a task should not only involve attributing mental states to others, but those should be different from their own (\emph{`non-merging' property} and an easier process than ToM should not be able to account for success in a particular task \emph{``mentalizing'' property}.

Those two principles show themselves in many different tasks, such as understanding pretend play, bluffing, white lies, etc., but one particular type of task became the litmus test of the field: the false belief test. Such a test involves understanding that an agent, when operating under incorrect information, will act according to it and not to the real state of the world \cite{schlingerTheoryMindOverview2009}. 

With the popularization of LLMs, it became key for those systems to have the necessary skills to engage in social interactions. In this context, researchers started discussing the extent to which they possess ToM. \cite{duijnTheoryMindLarge2023} showed that, as of 2023, most LLMs operated below the performance of children aged 7-10 in some LLM tasks. \cite{streetLLMsAchieveAdult2024} demonstrated that GPT-4 and Flan-PaLM achieve adult or near-adult performance in higher-order ToM tasks.

However, there is ongoing debate about whether the success of LLMs in those ToM tasks actually signifies that they truly developed ToM skills \cite{shapira_clever_2024}. \cite{ullmanLargeLanguageModels2023} demonstrates that, although GPT-3 passes ToM tasks,  alterations in prompts that seem trivial to us cause it to fail on them. The author claims that this means that GPT3's understanding of the underlying principle behind ToM is not real.

In the context of this paper, the concept of a real skill for AI regards simply what it shows via our interactions with it, so acting \textit{as if} it had a skill or \textit{really} having the skill are treated as equivalent. This is the case as claims about the true nature of AI's phenomenology and the possibility of it being a philosophical zombie, although valid and interesting discussions, are well beyond the scope of this paper, so Turing's more instrumentalist approach is adopted \cite{turing_icomputing_1950}.  It is important, however, to point out that it still makes sense to make claims like the one from \cite{ullmanLargeLanguageModels2023}, as the lack of `reality' in the AI understanding presents itself through properties of the model's response. 

Despite the discussion surrounding the validity of using the frequency of success in ToM tasks as a real assessment of a model having ToM skills, several benchmarks started to appear, each with a specific approach in the composition of the tasks.\cite{kimFANToMBenchmarkStresstesting2023} 
With benchmarks in place, it became easy to test if any particular technique could improve models' performance in ToM tasks. The first thing tested was the impact of CoT prompting on the models. It was found to increase their performance at least in a few cases \cite{kimFANToMBenchmarkStresstesting2023}. Also, other authors proposed other approaches, which include but are not limited to:
\begin{itemize}
    \item \emph{Perspective taking} prompts models to first filter the context to only what characters know before answering the ToM task \cite{wilf-etal-2024-think}.
    \item \emph{DEL-ToM} is an approach that constitutes scaling the inference time of models through an approach grounded in dynamic epistemic logic \cite{wuDELToMInferenceTimeScaling2025a}.
    \item \emph{Thought Tracing} is an inference-time reasoning algorithm that keeps track and weights different hypotheses surrounding characters' mental state in ToM tasks \cite{kimHypothesisDrivenTheoryofMindReasoning2025}.
\end{itemize}

A useful concept to shine additional light on ToM tasks is the one of Orders of Reasoning. To illustrate what this concept means, imagine three characters, Adu, Lin, and Lily. If Adu thinks "Lin is cool", this is considered a 0th-degree reasoning, as he is simply thinking about Lin, not addressing his mental state. On the other hand, if he thinks "Lilly thinks that Lin is cool", that would be a 1st-degree reasoning. To additionally illustrate higher degree reasoning, consider that Adu thinks "4) Lily thinks that 3) I think that 2) she thinks the 1) Lin thinks that 0) she's cool". In short, the order of reasoning of a ToM task is recursively how many mental representations it requires. \cite{meijeringKnowWhatYou}

\subsection{Third-Party Benchmark Results}\label{benchResults}

Efforts have already been made to test reasoning and non-reasoning models directly on established ToM benchmarks. Most relevant to us, \cite{kimHypothesisDrivenTheoryofMindReasoning2025} report results for a range of models, several reasoning models alongside their non-reasoning peers, across the FANToM, BigToM, MMToM-QA, and ParaphrasedToMi benchmarks. We reproduce their numbers in Table~\ref{benchmarks} for reference. Note that these models are slightly older than the ones used in our own experiments.

Two patterns are worth highlighting, which we return to in Section~\ref{Discussion}: reasoning models tend to score above comparable non-reasoning models on most benchmarks, and the gap is largest on FANToM, where a model only scores if it answers every question of a given type correctly, so inconsistency is penalised heavily. Both observations are consistent with the robustness-based reading we develop through the paper, and we treat them as supporting external evidence.

\begin{table*}[tbp]
\caption{Third party ToM benchmark results for reference, from \cite{kimHypothesisDrivenTheoryofMindReasoning2025}}\label{benchmarks}
\setlength{\tabcolsep}{2pt}
\begin{center}
\begin{tabular}{c c|c|c}
Model & ParaphrasedToMi & FANToM & MMToM-QA \\ \hline
\begin{tabular}[t]{p{0.21\linewidth}}
   \\
   \\
   GPT-4o\\ \hline
   o1-preview\\
   o1-low\\
   o1-medium\\
   o1-high\\ \hline
   o1-mini \\
   o3-mini-low \\
   o3-mini-medium \\
   o3-mini-high \\ \hline
   Deepseek R1 \\ \hline
   Llama 3.3 \\
   Gemini 1.5 Pro \\
   Qwen 2.5 72B \\
   QwQ 32B \\
   preview\\
   
\end{tabular} &
\begin{tabular}[t]{p{0.07\linewidth} p{0.07\linewidth} p{0.07\linewidth}}
   Avg. & False & True \\ 
        & Belief & Belief \\\hline
   59.5 & 38.5 & 80.5 \\ \hline
   68.0 & 100.0 & 36.0 \\
   67.8 & 98.5 & 37.0 \\
   70.8 & 96.0 & 45.5 \\
   73.0 & 97.0 & 49.0\\ \hline
   60.0 & 60.0 & 60.0\\
   64.0 & 62.5 & 65.5\\
   64.5 & 90.5 & 38.5\\
   64.0 & 99.5 & 28.5\\ \hline
   68.3 & 77.0 & 59.5\\ \hline
   57.3 & 44.5 & 70.0\\
   54.8 & 43.5 & 66.0\\
   57.3 & 39.0 & 75.5\\
   58.8 & 27.5 & 90.0\\
\end{tabular} &

\begin{tabular}[t]{p{0.06\linewidth} p{0.08\linewidth} p{0.08\linewidth}}
   All & Ans & I. Acc \\ 
   Qs. & All Qs. & All Qs.\\ \hline
   11.1& 33.3& 32.7\\ \hline
   44.4& 66.7& 63.5\\
   28.3& 41.5& 51.0\\
   30.2& 41.5& 58.8\\
   37.9& 44.8& 63.0\\ \hline
   9.4& 41.5& 23.5\\
   0.0& 17.0& 11.8\\
   1.9& 26.6& 19.6\\
   1.9& 35.8& 33.3\\ \hline
   37.9& 62.1& 48.1\\ \hline
   0.0& 7.4& 0.0\\
   1.9& 7.5& 2.0\\
   0.0& 5.7& 2.0\\
   0.0& 0.0& 0.0\\
\end{tabular}&
\begin{tabular}[t]{p{0.05\linewidth} p{0.065\linewidth} p{0.06\linewidth}}
   All & Belief & Goal \\
    & & \\ \hline
   56.5& 74.5& 39.8\\ \hline
   70.4& 96.0& 51.0\\
   76.5& 96.1& 57.1\\
   76.0& 94.1& 60.2\\
   76.5& 95.1& 59.2\\ \hline
   60.0& 91.2& 27.6\\
   55.0& 71.6& 37.8\\
   69.0& 94.2& 42.9\\
   71.5& 97.1& 44.9\\ \hline
   49.0& 73.5& 24.5\\ \hline
   47.0& 50.0& 42.9\\
   50.0& 70.6& 28.6\\
   41.5& 51.0& 32.7\\
   51.5& 71.7& 29.6\\
\end{tabular}
\end{tabular}
\end{center}
\end{table*}

\section{Methods}\label{methods}

To put the results in context, enable reproducibility, and promote replicability, this section presents an overview of the models considered in the study (Section \ref{models}), the psychological tests conducted on the models, including high-level implementation details, evaluation metrics, and references to the code and data used in the experiments (Section \ref{psic}), and a discussion on scoring (Section \ref{scoring}).

We emphasize that this paper does not provide a controlled causal identification of the effect of RLVR training. The evaluated systems differ in vendor interfaces, prompting affordances, and access to intermediate reasoning traces, which makes perfectly controlled comparisons difficult. Accordingly, our contribution should be understood as an empirical behavioral analysis: across a range of ToM tests and prompt perturbations, reasoning-oriented models appear more stable under variation, a pattern consistent with a robustness-based interpretation of recent performance gains.

All prompting classes, raw and graded data, experiment scripts, CLI grading tools, analysis notebooks, and additional detail such as ToM and ToM task definitions, prompting details, and qualitative insight from responses are publicly available at \href{https://github.com/ianbdehaan/Evaluating_ToM_in_Reasoning_Models}{this GitHub repository}.

\subsection{Models}\label{models}

Differences in the training of the models and the decisions of their creating companies affect how several properties of models vary. Some are crucial for the viability and reproducibility of the experiments run. Those differences are listed in Table \ref{tab:model-capabilities}.

\begin{table}[tbp]
    \caption{Properties of reasoning models used in the experiments.}
    \label{tab:model-capabilities}
    \centering
    \begin{tabular}{l|c|c|c|c}
         & GPT-5 & Claude & R1 & Grok-3-mini \\
        Thinking & Always on & Can turn off &  Always on &  Always on \\
        \hline
        Returned Reasoning & Sometimes & Frequently & Complete & Complete \\
         & summary & summary & & \\
        \hline
        Configurable Temperature & No & No & Yes & Yes \\
        \hline
        Reasoning as input& Yes & Yes & No & No \\
    \end{tabular}
\end{table}

 Claude is the only model analyzed for which it is possible to turn its thinking off entirely; this allows for experiments with the (think) Claude to be reproduced without the thinking, with any differences suggesting the impact that thinking has on its performance, so this is done. For all the models that allow for temperature configuration (Table~\ref{tab:model-capabilities}), it is set to 0  to increase reproducibility. For models that can keep the reasoning in context (Table~\ref{tab:model-capabilities}), this was done when re-prompting them.
 
 Both GPT-5 and Claude return reasoning summaries that might be filtered by the public interface rather than their full reasoning tokens; however, in preliminary tests, a very big discrepancy in summary qualities was detected, with GPT-5 often filtering most of its reasoning, while Claude's summary remains highly useful in understanding its reasoning in all prompts tried. Due to that, GPT-5 has to be additionally prompted about the reasoning behind its answers.
Note that comparisons are not perfectly controlled due to API constraints; results should be interpreted as behavioral, not architectural, comparisons.

\subsection{Psychological Tests} \label{psic}

The psychological tests that are presented in \cite{duijnTheoryMindLarge2023} are applied to reasoning models to assess how well the current reasoning models perform in ToM tasks compared to those reported in the 2023 baseline reference. This includes classical and modified first and second-order Sally-Anne tests, a Strange Stories test, an Imposing Memory test, and modifications on simple ToM task tests. 
The latter test was based on the principles behind the prompt modifications shown to make GPT3 fail on simple ToM tasks \cite{ullmanLargeLanguageModels2023}. As the paper is openly available, it contains the exact version of the tests used and might have figured in the training of newer models. To avoid having models succeed due to their training data, a new task based on each principle was written. 
Full details on test and modifications can be found in the supplementary materials in the repository.

\subsection{Scoring}\label{scoring}

Models are evaluated with the same metric for scoring performance as the one used by \cite{duijnTheoryMindLarge2023}, with the exception that, for reasoning models that display it, the reasoning is used as the explanation rather than the answer to a follow-up question. In detail, 0 points if the reasoning is incorrect, 1 point if the reasoning comes close to the right answer, \textbf{or} if the model subsequently claims they were wrong (moving from or to a correct answer with proper explanation), and  2 points if the result and reasoning are correct. In the results section (\ref{Results}), all the reported scores are normalized to the range $[0,1]$ by simply dividing the average result by 2.

\section{Results} \label{Results}
In this section, we present the results from the psychological tests (Section~\ref{sallyAnne}--\ref{modPrompts}) and we share selected findings from qualitative inspection of model responses to provide insights into both successes and failures in ToM tasks (Section~\ref{furtherInsights}).

\subsection{1st and 2nd Degree Sally Anne}\label{sallyAnne}

\begin{table}[tbp]
    \caption{Results obtained for Sally-Anne (SA\_1fb) tests and second-degree Sally-Anne (SA\_2fb) tests for the different models. Results are averaged over samples.}
    \label{tab:Sally-Anne}
    \setlength{\tabcolsep}{8pt}
    \centering
    \input{Tables/SA_test}
\end{table}

As seen in Table \ref{tab:Sally-Anne}, almost all the answers were correct and had a proper reasoning associated with them. The only exception is a case where grok-3-mini ignores a character's mental state and merges the knowledge of the two characters when reasoning about the question, merging two characters' knowledge and, therefore, arriving at the wrong answer. These results show substantially better performance in these models than the ones verified a few years ago by \cite{duijnTheoryMindLarge2023}.

 Even though for each particular degree of reasoning three different stories are presented, with two being modifications from the original Sally-Anne test, the good performance in this task alone might still be attributed to the widespread knowledge about this psychological test, as it is the most famous interpretation of the false-belief test and the general semantic structure of the three versions of the test remain the same, possibly allowing a model to get to the answer through spurious correlations.
 The performance in this particular test is too high for any extra analysis.
 
\subsection{Strange Stories}\label{strangeStories}

\begin{table}[tbp]
    \caption{Results obtained for the seven types of Strange Stories for different models. Results are averaged over samples.}
    \label{tab:Strange Stories}
    \setlength{\tabcolsep}{1pt}
    \centering
    \input{Tables/SS_test_local}
\end{table}

The performance of the models in the first five strange stories categories, which involves  understanding of, respectively, lies, pretend play, jokes, white lies, and misunderstanding, was flawless (Table \ref{tab:Strange Stories}).
Some models showed slightly more difficulty in the last two categories, sarcasm and double bluff (Table \ref{tab:Strange Stories}). With sarcasm, both Claude and Grok-3-mini made partial errors in the same story. The story revolves around a father asking his son to clean the kitchen. The son decides to make an extra effort and also clean the inside of the cabinets, but when he opens it, a pack of flour falls and explodes. Then, the father gets in the kitchen and says, ``Wow, everything is so clean now!''. Both models considered, in their reasoning, that the father might have been sarcastic, but ultimately decide he simply looked at the part of the kitchen that was already cleaned and still had not seen the mess, which makes some sense given that the models can't draw a mental representation of the kitchen to understand it would be almost impossible not to see the mess.

In the double bluff case, the same thing happens: the three models, in the same story, consider double bluff as a possible explanation in their reasoning, but end up going with another explanation that still makes some logical sense. The story revolves around a hide-and-seek game, where the person playing hide is described as very smart. The person is trying to avoid being found immediately. They consider a shed, which offers better concealment, and a tree as the possible hiding spots; they consider that the shed is the obvious choice, but end up going with it anyway. The models that respond incorrectly consider they might have been double bluffing, but ultimately decide the key is not to be found \emph{immediately}, so the superior concealment of the shed is enough.

The performance of the models as a whole is close to perfect in the Strange Stories task, indicating how much better models have gotten with time. Specifically GPT-5 properly answered and justified all questions. Interestingly, the two cases in which models partly failed could definitely use a mental visual depiction of a situation to make it clearer. This will be further discussed in the discussion (Section~\ref{Discussion}). Lastly, it is interesting to note that, in this task, Claude, with thinking off performed as well as its reasoning counterpart,  but the error pattern is different: Claude is better on sarcasm, worse on double bluff. It is speculated that the overall high performance is because the strange stories are pretty straightforward, so the possible solution paths are probably not that diverse.

\subsection{Imposing Memories}\label{impMemories}

\begin{table*}[tbp]
    \caption{Results obtained for the Imposing Memories' tasks for different models. Results are averaged over samples.}
    \label{tab:Imposing memories}
    \centering
    \setlength{\tabcolsep}{3pt}
    \begin{tabular}{c|c}
    Intentionality & Memory \\ \hline
    \input{Tables/HM_i_test} & \input{Tables/HM_m_test}  \\
    \end{tabular}
\end{table*}

The models performed really well in the stories overall, with Claude (thinking on) responding correctly to all the questions (Table~\ref{tab:Imposing memories}). The types of mistakes made vary, but include:

\begin{itemize}
    \item Inferring things that are not in the story -- i.e., he wanted to buy a post for sending a card to his grandmother $\rightarrow$ he has to buy a card for his grandmother
    \item Not understanding that the question is about a character's mental state rather than reality  -- i.e., for the  statement 'Hannah: I thought Abi went home sick' thinking 'Abi went home sick, so it is True'
    \item Failing to understand higher degree ToM -- i.e. for statement 'Hannah: I thought Ama knew that Abi had gone home sick. Is this true?' responding 'No, it is not true that Ama knew Abi had gone home sick(...)'
    \item Hallucinating and responding to something else entirely
\end{itemize}

The thinking version of Claude performed slightly better than its non-thinking counterpart (100\% vs 92\%). We hypothesize that the difference is more pronounced in this test than in the last two as the inference-time scaling provided by the reasoning is probably very useful while parsing the story details carefully and comparing them with the claim.

This task's difficulty revolves a lot around recovering facts from the histories, which is why it is useful to compare the performance of models in the intentionality questions with the memory ones. Models made a bit more than twice as many errors in the intentionality class as in the memory one. However, the errors in the intentionality class were made in the same questions, making the fact that those were hard questions also a valid explanation for the discrepancy. With this in mind, it is impossible to draw further conclusions from this test.

\subsection{Modifications on simple prompts}\label{modPrompts}

\begin{table}[tbp]
    \caption{Performance of the Models over modified tasks.}
    \label{tab:modified tasks}
    \setlength{\tabcolsep}{0.8pt}
    \centering
    \input{Tables/mod_test}
\end{table}

The models we tested perform better on these modifications than GPT-3 did in \cite{ullmanLargeLanguageModels2023}, suggesting they are more robust on this specific task than GPT-3 was (see supplementary materials in repository for exact definition of modifications). Two of these tasks, 2A and 2B, are nonetheless still particularly difficult for them (see Table~\ref{tab:modified tasks}). In the next section, we hypothesize that what makes them hard is that the modifications are only trivial when one can picture the scene mentally, and solving them semantically would rely on specifically crafted heuristics.

Additionally, it's possible to verify that the reasoning version of Claude performed significantly better in these tests than its non-thinking counterpart. It is hypothesized that, with prompts designed to distract models from the optimal reasoning, the model's reasoning process makes a substantial difference to performance.

\subsection{Qualitative Insights from Reasoning Responses}
\label{furtherInsights}

Qualitative inspection of model responses reveals several recurring reasoning patterns that help explain both successes and failures in ToM tasks. 

In many cases, models begin by explicitly filtering and restructuring the narrative into a sequence of factual observations, which appears to help track state changes and identify relevant information for the task. Reasoning-oriented models also sometimes perform a perspective-taking step without explicit prompting, separating what different characters know or believe in a manner similar to perspective-taking prompting strategies proposed in prior work. At the same time, reasoning traces occasionally reveal meta-knowledge about the task itself, such as identifying a scenario as a `classic false-belief test,' which could in principle serve as a heuristic for solving certain tasks through pattern recognition rather than genuine mental-state reasoning. However, it seems unlikely that meta-knowledge alone accounts for the broader gains, since tasks like the modifications don't match a known template, though we can't rule this out quantitatively 

Several failure modes also appear in the traces, including indecision between multiple plausible interpretations, confusion between story content and the question being asked, and incorrect reasoning paths that can sometimes be corrected when the model is prompted to reconsider its answer. Taken together, these qualitative observations suggest that many errors arise from instability in selecting the correct inference path rather than from a complete absence of task-relevant knowledge, a pattern consistent with our broader interpretation that recent improvements in ToM task performance are associated with increased robustness rather than fundamentally new reasoning capabilities. 
For a range of response examples illustrating the above, see the supplementary materials.

\section{Discussion}
\label{Discussion}
In this section, we reflect on the impact of the results given the literature (Section~\ref{reflection}) and discuss known limitations and future research (Section~\ref{limitations}).

\subsection{Reflection on Results}\label{reflection}

The results show that models have improved substantially in ToM tasks since 2023. Part of this is likely just that models have become more capable overall, but in what follows we focus on a second factor: a newfound robustness in their ToM skills.

Yue et al. \cite{yueDoesReinforcementLearning2025} show that the reasoning paths produced by reasoning models stay bounded by their base models, reaching asymptotic performance without lowering perplexity on the base model, reasoning training does not appear to add fundamentally new capabilities. Read alongside our own results, this points to a specific interpretation. In our experiments the thinking-enabled models, and most clearly the thinking-on version of Claude compared to its thinking-off counterpart, were more robust to prompt and task variation. We therefore hypothesize that the main effect of this kind of reasoning training is improved stability in reaching a solution the model could already, in principle, reach, rather than an expansion of representational capacity. We stress that our setup compares thinking and non-thinking models rather than RLVR-trained and non-RLVR-trained ones, so we cannot attribute this effect to RLVR specifically (see Section~\ref{limitations}).

The psychological tests show just how robust the new models can be: GPT-5, for instance, made only a single partial mistake across all the tests taken from \cite{duijnTheoryMindLarge2023}. This robustness seems to matter most in tests specifically designed to disturb a model's reasoning, which is where the gap between thinking and non-thinking models was widest, as discussed next.

Two of the `modification' tasks weren't successfully completed by a single model. This may be because those modifications are not as trivial as one might have anticipated, as they might require picturing the scene: for humans, picturing scenes in our minds is so common that it is natural to forget how essential of a skill it can be. Of course, one can still, through many steps, derive the consequences of transparency semantically for instance, but it becomes a substantially harder task than anticipated.
It is also interesting to notice that one of the only two stories in the 'Strange Stories' psychological test in which models made a partial mistake had a visual aspect to it that could clarify which of the two hypotheses the models considered in their reasoning is the correct one. 

As seen in Section~\ref{ToM and LLMs}, various prompting strategies have emerged over the years to push LLMs to perform better on ToM tasks. This suggests models often already had the necessary skills but needed the right prompt to surface them reliably. Our results point the same way: the models sometimes carry out a version of what perspective-taking prompting aims to induce without being prompted to do so.

This is also in line with the findings of \cite{ullmanLargeLanguageModels2023}, which showed that simple prompt alterations would lead the models to fail in the same types of tasks. Here we conjecture that this means that the upper bound for success was already present in the models, not that they didn't possess 'real' ToM skills. In other words, they already, to a large extent, had the necessary skills to solve ToM problems, but weird prompts could easily disturb their reasoning.

To corroborate this, we analyzed the benchmark results from \cite{kimHypothesisDrivenTheoryofMindReasoning2025}, which proposes a Bayesian theory-of-mind framework aligned with our robustness hypothesis. Although featuring slightly older reasoning models, the results show they consistently outperform their non-reasoning counterparts on ToM tasks, with even larger gains on the benchmark that most strongly penalizes a lack of robustness.

With this in mind, the result from \cite{yueDoesReinforcementLearning2025} is hereby reinterpreted, at least in the context of ToM tasks. By understanding what reasoning models don't do, it is also possible to understand what they do. More broadly, we hypothesize that the narrowing of the reasoning coverage of RLVR models might be precisely their strength. While one might want models with very flexible results for some applications, it's generally desirable for models to be robust.

\subsection{Limitations and Further Research}\label{limitations}

Although evidence from our experiments and benchmarks in literature were found of reasoning helping models in answering queries more robustly in the context of ToM tasks, no clear quantification of such an effect can be provided with only this work. Future research could tackle such a problem by running ToM benchmark tests comparing RLVR-trained models with their base model counterparts using the pass@k metric proposed by \cite{yueDoesReinforcementLearning2025}.  

Additionally, spurious correlations like relating the semantic structure of the tests with meta-knowledge of ToM can't be ruled out for all psychological test cases, which might make some of the results inconclusive for real-world scenarios.

The present work also does not directly compare the RLVR-trained models' performance in ToM with base models employing other inference-time scaling techniques; the extent to which the increase in performance can be attributed directly to this specific way of increasing inference time can be investigated in future work.

Furthermore, some of the evidence motivating the conclusions of this paper is drawn by comparing the performance of Claude with thinking on with its counterpart with thinking off. Claude, with thinking off is not the same thing as its base model, which is not accessible, yet we hope the difference in performance sheds some light on what the reasoning process is actually helping the model with. Further research could investigate differences in performance between reasoning models and base models in cases where they are available, such as R1 and V3-base.
It's also noteworthy that, although the reasoning traces were analyzed for strategies employed by the models, it has been argued that reasoning traces don't always present the real motivation behind a model's final response \cite{chenReasoningModelsDont2025}. 

Lastly, the different treatment between models might induce differences in performance. For example, setting the models' temperature to 0 or asking why a response was given directly affects the performance of the models in the given tests.

\section{Conclusion}\label{conclusions}

This paper examines the behavior of recent reasoning-oriented language models on Theory of Mind tasks using a combination of psychological tests, prompt perturbations, qualitative analysis of responses, and benchmark results from prior work. Across these analyses, reasoning-oriented models appeared consistently more robust to variations in prompt form and task presentation. We read this as evidence that recent gains in ToM evaluations reflect improvements in robustness and inference stability, rather than taking it as proof of new ToM-specific reasoning capabilities, which our behavioral setup cannot directly test.

More broadly, our results suggest that progress on ToM tasks should not be assessed only through average benchmark performance, but also through robustness under task-preserving variations. Given this, some earlier prompt-sensitive failures may be better understood as failures to reliably reach the correct inference path rather than evidence of a complete absence of task-relevant competence. At the same time, our comparisons are behavioral rather than fully controlled, so the conclusions should be interpreted as evidence consistent with a robustness-based account rather than as a causal attribution to RLVR training.

Overall, the paper highlights two broader points: first, current models perform substantially better on ToM tasks than earlier generations; second, robust\-ness-oriented evaluation is essential for interpreting apparent social-cognitive progress in LLMs. We hope these findings contribute to a more careful understanding of what reasoning-oriented models improve, and to the design of future evaluations that better distinguish robustness from underlying capability.

\begin{credits}

\subsubsection{\discintname}

The authors have no competing interests to declare that are relevant to the content of this article. 

\end{credits}

%
%
\bibliographystyle{splncs04}
\bibliography{bibfile}
%

\end{document}

%% file: Tables/SA_test.tex
\begin{tabular}{l|c|c|c|c|c}
 & Claude & R1 & Claude-no- & GPT-5- & Grok-3- \\
 &  &  & thinking & high & mini \\
 \hline

 SA-1 & \cellcolor{yellowcell}\textbf{1.00} & \cellcolor{yellowcell}\textbf{1.00} & \cellcolor{yellowcell}\textbf{1.00} & \cellcolor{yellowcell}\textbf{1.00} & \cellcolor{yellowcell}\textbf{1.00} \\
  SA-2 & \cellcolor{yellowcell}\textbf{1.00} & \cellcolor{yellowcell}\textbf{1.00} & \cellcolor{yellowcell}\textbf{1.00} & \cellcolor{yellowcell}\textbf{1.00} & \cellcolor{darkorangecell}\textcolor{lighttext}{0.67} \\\hline
\end{tabular}

%% file: Tables/SS_test_local.tex
\begin{tabular}{l c|c|c|c|c|c|c}
 & lie & pretend & joke & whitelie & misunderstanding & sarcasm & double-bluff \\
\hline
claude & \cellcolor{yellowcell}\textcolor{black}{\textbf{1.00}} & \cellcolor{yellowcell}\textcolor{black}{\textbf{1.00}} & \cellcolor{yellowcell}\textcolor{black}{\textbf{1.00}} & \cellcolor{yellowcell}\textcolor{black}{\textbf{1.00}} & \cellcolor{yellowcell}\textcolor{black}{\textbf{1.00}} & \cellcolor{lightorangecell}\textcolor{black}{\textbf{0.83}} & \cellcolor{lightorangecell}\textcolor{black}{\textbf{0.83}} \\
r1 & \cellcolor{yellowcell}\textcolor{black}{\textbf{1.00}} & \cellcolor{yellowcell}\textcolor{black}{\textbf{1.00}} & \cellcolor{yellowcell}\textcolor{black}{\textbf{1.00}} & \cellcolor{yellowcell}\textcolor{black}{\textbf{1.00}} & \cellcolor{yellowcell}\textcolor{black}{\textbf{1.00}} & \cellcolor{yellowcell}\textcolor{black}{\textbf{1.00}} & \cellcolor{lightorangecell}\textcolor{black}{\textbf{0.83}} \\
gpt-5-high & \cellcolor{yellowcell}\textcolor{black}{\textbf{1.00}} & \cellcolor{yellowcell}\textcolor{black}{\textbf{1.00}} & \cellcolor{yellowcell}\textcolor{black}{\textbf{1.00}} & \cellcolor{yellowcell}\textcolor{black}{\textbf{1.00}} & \cellcolor{yellowcell}\textcolor{black}{\textbf{1.00}} & \cellcolor{yellowcell}\textcolor{black}{\textbf{1.00}} & \cellcolor{yellowcell}\textcolor{black}{\textbf{1.00}} \\
grok-3-mini & \cellcolor{yellowcell}\textcolor{black}{\textbf{1.00}} & \cellcolor{yellowcell}\textcolor{black}{\textbf{1.00}} & \cellcolor{yellowcell}\textcolor{black}{\textbf{1.00}} & \cellcolor{yellowcell}\textcolor{black}{\textbf{1.00}} & \cellcolor{yellowcell}\textcolor{black}{\textbf{1.00}} & \cellcolor{lightorangecell}\textcolor{black}{\textbf{0.83}} & \cellcolor{yellowcell}\textcolor{black}{\textbf{1.00}} \\
claude-no- & \cellcolor{yellowcell}\textcolor{black}{\textbf{1.00}} & \cellcolor{yellowcell}\textcolor{black}{\textbf{1.00}} & \cellcolor{yellowcell}\textcolor{black}{\textbf{1.00}} & \cellcolor{yellowcell}\textcolor{black}{\textbf{1.00}} & \cellcolor{yellowcell}\textcolor{black}{\textbf{1.00}} & \cellcolor{yellowcell}\textcolor{black}{\textbf{1.00}} & \cellcolor{darkorangecell}\textcolor{lighttext}{\textbf{0.67}} \\
thinking &\cellcolor{yellowcell} &\cellcolor{yellowcell} &\cellcolor{yellowcell} &\cellcolor{yellowcell} & \cellcolor{yellowcell} & \cellcolor{yellowcell} & \cellcolor{darkorangecell} \\
\hline
\end{tabular}

%% file: Tables/HM_i_test.tex
\begin{tabular}{lrrrr}
 & 2 & 3 & 4 & 5 \\
\hline
claude & \cellcolor{yellowcell}\textcolor{black}{\textbf{1.00}} & \cellcolor{yellowcell}\textcolor{black}{\textbf{1.00}} & \cellcolor{yellowcell}\textcolor{black}{\textbf{1.00}} & \cellcolor{yellowcell}\textcolor{black}{\textbf{1.00}} \\
r1 & \cellcolor{yellowcell}\textcolor{black}{\textbf{1.00}} & \cellcolor{redcell}\textcolor{lighttext}{\textbf{0.50}} & \cellcolor{yellowcell}\textcolor{black}{\textbf{1.00}} & \cellcolor{yellowcell}\textcolor{black}{\textbf{1.00}} \\
claude-no-thinking & \cellcolor{yellowcell}\textcolor{black}{\textbf{1.00}} & \cellcolor{redcell}\textcolor{lighttext}{\textbf{0.50}} & \cellcolor{yellowcell}\textcolor{black}{\textbf{1.00}} & \cellcolor{yellowcell}\textcolor{black}{\textbf{1.00}} \\
gpt-5-high & \cellcolor{yellowcell}\textcolor{black}{\textbf{1.00}} & \cellcolor{orangecell}\textcolor{lighttext}{\textbf{0.75}} & \cellcolor{yellowcell}\textcolor{black}{\textbf{1.00}} & \cellcolor{yellowcell}\textcolor{black}{\textbf{1.00}} \\
grok-3-mini & \cellcolor{yellowcell}\textcolor{black}{\textbf{1.00}} & \cellcolor{orangecell}\textcolor{lighttext}{\textbf{0.75}} & \cellcolor{yellowcell}\textcolor{black}{\textbf{1.00}} & \cellcolor{yellowcell}\textcolor{black}{\textbf{1.00}} \\
\hline
\end{tabular}

%% file: Tables/HM_m_test.tex
\begin{tabular}{rrrrr}
 1 & 2 & 3 & 4 & 5 \\
\hline
\cellcolor{yellowcell}\textcolor{black}{\textbf{1.00}} & \cellcolor{yellowcell}\textcolor{black}{\textbf{1.00}} & \cellcolor{yellowcell}\textcolor{black}{\textbf{1.00}} & \cellcolor{yellowcell}\textcolor{black}{\textbf{1.00}} & \cellcolor{yellowcell}\textcolor{black}{\textbf{1.00}} \\
\cellcolor{yellowcell}\textcolor{black}{\textbf{1.00}} & \cellcolor{yellowcell}\textcolor{black}{\textbf{1.00}} & \cellcolor{yellowcell}\textcolor{black}{\textbf{1.00}} & \cellcolor{orangecell}\textcolor{lighttext}{\textbf{0.75}} & \cellcolor{yellowcell}\textcolor{black}{\textbf{1.00}} \\
\cellcolor{lightorangecell}\textcolor{black}{\textbf{0.88}} & \cellcolor{yellowcell}\textcolor{black}{\textbf{1.00}} & \cellcolor{yellowcell}\textcolor{black}{\textbf{1.00}} & \cellcolor{lightorangecell}\textcolor{black}{\textbf{0.88}} & \cellcolor{yellowcell}\textcolor{black}{\textbf{1.00}} \\
\cellcolor{yellowcell}\textcolor{black}{\textbf{1.00}} & \cellcolor{yellowcell}\textcolor{black}{\textbf{1.00}} & \cellcolor{yellowcell}\textcolor{black}{\textbf{1.00}} & \cellcolor{yellowcell}\textcolor{black}{\textbf{1.00}} & \cellcolor{yellowcell}\textcolor{black}{\textbf{1.00}} \\
\cellcolor{yellowcell}\textcolor{black}{\textbf{1.00}} & \cellcolor{yellowcell}\textcolor{black}{\textbf{1.00}} & \cellcolor{yellowcell}\textcolor{black}{\textbf{1.00}} & \cellcolor{orangecell}\textcolor{lighttext}{\textbf{0.75}} & \cellcolor{yellowcell}\textcolor{black}{\textbf{1.00}} \\
\hline
\end{tabular}

%% file: Tables/mod_test.tex
\begin{tabular}{lc|c|c|c|c|c|c|c|c|c|c|c|c|c}
 & 1A.1 & 1A.2 & 1B.1 & 1B.2 & 1C.1 & 1C.2 & 1D.1 & 1D.2 & 2A.1 & 2A.2 & 2B.1 & 2B.2 & 2C.1 & 2C.2 \\
\hline
claude & \cellcolor{yellowcell}\textcolor{black}{1.0} & \cellcolor{yellowcell}\textcolor{black}{1.0} & \cellcolor{yellowcell}\textcolor{black}{1.0} & \cellcolor{yellowcell}\textcolor{black}{1.0} & \cellcolor{yellowcell}\textcolor{black}{1.0} & \cellcolor{yellowcell}\textcolor{black}{1.0} & \cellcolor{yellowcell}\textcolor{black}{1.0} & \cellcolor{yellowcell}\textcolor{black}{1.0} & \cellcolor{blackcell}\textcolor{lighttext}{0.0} & \cellcolor{blackcell}\textcolor{lighttext}{0.0} & \cellcolor{blackcell}\textcolor{lighttext}{0.0} & \cellcolor{blackcell}\textcolor{lighttext}{0.0} & \cellcolor{yellowcell}\textcolor{black}{1.0} & \cellcolor{yellowcell}\textcolor{black}{1.0} \\
grok-3-mini & \cellcolor{yellowcell}\textcolor{black}{1.0} & \cellcolor{redcell}\textcolor{lighttext}{0.5} & \cellcolor{yellowcell}\textcolor{black}{1.0} & \cellcolor{yellowcell}\textcolor{black}{1.0} & \cellcolor{yellowcell}\textcolor{black}{1.0} & \cellcolor{yellowcell}\textcolor{black}{1.0} & \cellcolor{yellowcell}\textcolor{black}{1.0} & \cellcolor{redcell}\textcolor{lighttext}{0.5} & \cellcolor{blackcell}\textcolor{lighttext}{0.0} & \cellcolor{blackcell}\textcolor{lighttext}{0.0} & \cellcolor{blackcell}\textcolor{lighttext}{0.0} & \cellcolor{blackcell}\textcolor{lighttext}{0.0} & \cellcolor{yellowcell}\textcolor{black}{1.0} & \cellcolor{yellowcell}\textcolor{black}{1.0} \\
gpt-5-high & \cellcolor{yellowcell}\textcolor{black}{1.0} & \cellcolor{yellowcell}\textcolor{black}{1.0} & \cellcolor{yellowcell}\textcolor{black}{1.0} & \cellcolor{blackcell}\textcolor{lighttext}{0.0} & \cellcolor{yellowcell}\textcolor{black}{1.0} & \cellcolor{yellowcell}\textcolor{black}{1.0} & \cellcolor{yellowcell}\textcolor{black}{1.0} & \cellcolor{redcell}\textcolor{lighttext}{0.5} & \cellcolor{redcell}\textcolor{lighttext}{0.5} & \cellcolor{redcell}\textcolor{lighttext}{0.5} & \cellcolor{blackcell}\textcolor{lighttext}{0.0} & \cellcolor{blackcell}\textcolor{lighttext}{0.0} & \cellcolor{yellowcell}\textcolor{black}{1.0} & \cellcolor{yellowcell}\textcolor{black}{1.0} \\
r1 & \cellcolor{yellowcell}\textcolor{black}{1.0} & \cellcolor{yellowcell}\textcolor{black}{1.0} & \cellcolor{yellowcell}\textcolor{black}{1.0} & \cellcolor{redcell}\textcolor{lighttext}{0.5} & \cellcolor{redcell}\textcolor{lighttext}{0.5} & \cellcolor{yellowcell}\textcolor{black}{1.0} & \cellcolor{yellowcell}\textcolor{black}{1.0} & \cellcolor{yellowcell}\textcolor{black}{1.0} & \cellcolor{blackcell}\textcolor{lighttext}{0.0} & \cellcolor{blackcell}\textcolor{lighttext}{0.0} & \cellcolor{blackcell}\textcolor{lighttext}{0.0} & \cellcolor{blackcell}\textcolor{lighttext}{0.0} & \cellcolor{yellowcell}\textcolor{black}{1.0} & \cellcolor{yellowcell}\textcolor{black}{1.0} \\ \hline
no-reas-claude & \cellcolor{redcell}\textcolor{lighttext}{0.5} & \cellcolor{redcell}\textcolor{lighttext}{0.5} & \cellcolor{yellowcell}\textcolor{black}{1.0} & \cellcolor{redcell}\textcolor{lighttext}{0.5} & \cellcolor{yellowcell}\textcolor{black}{1.0} & \cellcolor{yellowcell}\textcolor{black}{1.0} & \cellcolor{redcell}\textcolor{lighttext}{0.5} & \cellcolor{redcell}\textcolor{lighttext}{0.5} & \cellcolor{blackcell}\textcolor{lighttext}{0.0} & \cellcolor{blackcell}\textcolor{lighttext}{0.0} & \cellcolor{blackcell}\textcolor{lighttext}{0.0} & \cellcolor{blackcell}\textcolor{lighttext}{0.0} & \cellcolor{redcell}\textcolor{lighttext}{0.5} & \cellcolor{yellowcell}\textcolor{black}{1.0} \\
\hline
\end{tabular}